\documentclass[a4paper]{svproc}
\usepackage{url}

\usepackage[numbers]{natbib}
\usepackage{subcaption}
\usepackage{amsmath}
\usepackage{mathabx}
\usepackage[bookmarks=true]{hyperref}
\usepackage{graphicx}
\DeclareMathOperator*{\argmin}{arg\, min}

\renewcommand\vec{\mathbf}

\begin{document}
\mainmatter             
\title{Combining Coarse and Fine Physics for Manipulation using Parallel-in-Time Integration}

\titlerunning{Combining Coarse and Fine Physics for Manipulation} 

\institute{School of Computing, University of Leeds,  United Kingdom.\\
\and
School of Mechanical Engineering, University of Leeds,  United Kingdom.}

\author{Wisdom C. Agboh\inst{1} \and Daniel Ruprecht\inst{2} \and Mehmet R. Dogar\inst{1}}
\authorrunning{Agboh et al.}
\maketitle             
\begin{abstract}
We  present  a  method  for  fast  and  accurate  physics-based predictions  during  non-prehensile  manipulation  planning and control. Given  an  initial  state  and  a  sequence  of  controls,  the problem of predicting the resulting sequence of states is a key component  of  a  variety  of  model-based  planning  and  control algorithms. We propose combining a coarse (i.e. computationally cheap but not very accurate) predictive physics model, with a fine (i.e. computationally expensive but accurate) predictive physics model, to  generate  a  hybrid  model  that  is  at  the  required  speed  and accuracy for a given manipulation task.
Our approach is based on the Parareal algorithm, a parallel-in-time  integration  method  used  for  computing  numerical solutions for general systems of ordinary differential equations.
We adapt Parareal to combine a coarse pushing model with an off-the-shelf physics engine to  deliver physics-based predictions  that  are  as  accurate  as  the  physics engine  but  run  in substantially less wall-clock time, thanks to parallelization across time. We use these physics-based predictions in a model-predictive-control framework based on trajectory optimization, to plan pushing actions that avoid an obstacle and reach a goal location.
We show that with hybrid physics models, we can achieve the same success rates as the planner that uses the off-the-shelf physics engine directly, but significantly faster. We present experiments in simulation and on a real robotic setup. Videos are available here: \url{https://youtu.be/5e9oTeu4JOU}.  
\vspace{-2mm}
\end{abstract}
\section{Introduction}
\begin{figure}[t]
\centering 
\captionsetup{justification=centering}
  \begin{subfigure}[b]{0.245\textwidth}
  \centering
  \hspace{10mm}
    \includegraphics[scale=0.22, angle=0]{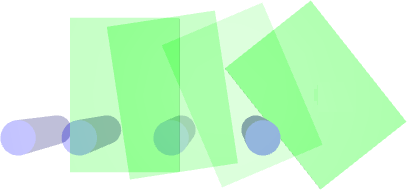}
    \caption{Coarse physics model (7 ms)}
    \label{fig:coarse_object_trajectory}
  \end{subfigure}
    \begin{subfigure}[b]{0.24\textwidth}
    \centering
    \includegraphics[scale=0.22, angle=0]{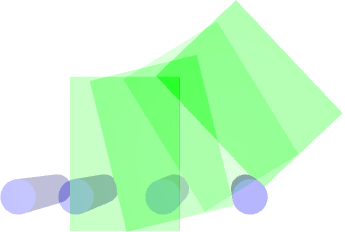}
    \caption{Hybrid physics model-1 (168 ms)}
    \label{fig:parareal_1_object_trajectory}
  \end{subfigure}
    \begin{subfigure}[b]{0.24\textwidth}
    \centering
    \includegraphics[scale=0.22, angle=0]{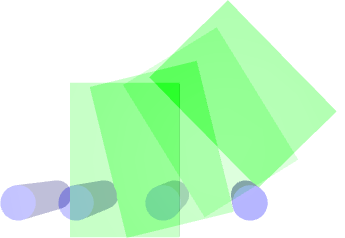}
    \caption{Hybrid physics model-2 (350 ms)}
    \label{fig:parareal_2_object_trajectory}
  \end{subfigure}
    \begin{subfigure}[b]{0.24\textwidth}
    \centering
    \includegraphics[scale=0.22, angle=0]{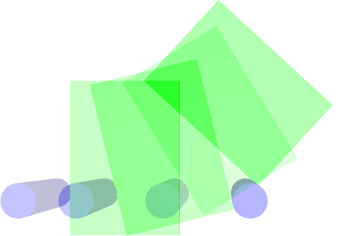}
    \caption{Physics engine prediction (668 ms)}
    \label{fig:fine_physics_object trajectory}
  \end{subfigure}
  \caption{A spectrum of physics predictions ranging from cheapest and least accurate (a) to expensive and most accurate (d).}
  \label{fig:spectrum_of_physics_models}
   \vspace{-4mm}
 \end{figure}

\vspace{-1mm}
We present a method for fast and accurate physics predictions
during non-prehensile manipulation planning and control.
Take the case study in Fig.\ref{fig:spectrum_of_physics_models}, where a
cylindrical object moves towards the right, pushing a box. We are interested in
predicting the motion of the pushed box, in a fast and accurate way.
To achieve this, we combine \textit{coarse} physics models with
\textit{fine} physics models.
By coarse models, we
mean computationally cheap but relatively inaccurate predictive physical models.  For example in
Fig.\ref{fig:coarse_object_trajectory}, we use a coarse model to compute the
motion of the box. The motion
is not completely realistic, but we can compute it extremely fast (7 ms wall-clock time to compute a simulated 8 s push).  By fine
models, we mean computationally expensive but accurate predictive physical models.  In
Fig.\ref{fig:fine_physics_object trajectory}, we use a fine model (in this case, the
Mujoco simulator \citep{mujoco}) to compute the motion of the same box. The
motion is more realistic, but it also requires much more time to compute (668
ms). 

We combine these two models to deliver a prediction that is as accurate as the fine model but runs in substantially less wall-clock time.
The motion predicted in Fig.\ref{fig:parareal_1_object_trajectory} is similar to the fine model prediction, but is four times faster to compute. 
The motion predicted in Fig.\ref{fig:parareal_2_object_trajectory} is indistinguishable from the fine model prediction for real world manipulation purposes, and is two times faster to compute.

Given an initial state and a sequence of controls, the problem of predicting
the resulting sequence of states is a key component of a variety of model-based
planning and control algorithms
\citep{pusher_slider,kalakrishnan2011stomp,convergent_planning,mppi,mppi_push,haustein_asfour,real_time_manipulation_Agboh_Humanoids18,tassa_synthesis,ilqr}.
Mathematically, such a prediction requires solving an \textit{initial value
problem}. 
Typically, those are solved through  numerical integration over time-steps (e.g. Euler's method or Runge-Kutta methods) using an underlying physics model.  
However, the speed with which these accurate physics-based predictions can be performed is still slow \citep{compare_engine_ErezICRA2015} and faster physics-based predictions can contribute significantly to
contact-based/non-prehensile manipulation planning and control.

There are several ways that could be used to construct coarse models for manipulation planning.
Quasi-static physics, which ignores accelerations, is widely used in
non-prehensile manipulation planning and control
\citep{coarse_model_Lynch_IROS92,mechanics_mason_ijrr86,planar_sliding_with_dry_friction_goyal_wear91},
and can be seen as a coarse model. Learning is another method which can be used to generate approximate but fast predictions \citep{learned_pushing_Mericli_AutoRob15,learned_pushing_Kopicki_AutoRob17,effect_aware_push,convex_polynomial_model_Zhou_IJRR18}.
Recently, with the advance of deep-learning, there has been multiple attempts at learning approximate ``intuitive'' physics models which are then used for manipulation planning
\citep{agrawal2016learning,finn2016unsupervised,finn2017deep,Rajeswaran-RSS-18,Matas-CORL2018,Ebert-CORL2018,augmenting_physics_Ajay_IROS18}. Especially when these networks are faced with novel objects that are not in their training data (e.g. consider a network trained with boxes and cylinders, but used to predict the motion of an ellipse) they can generate approximate predictions of motion, and therefore are good candidates as coarse models.

A key question we investigate is whether we can combine such cheap but approximate models, with expensive but more accurate and general models (such as physics engines) to generate a hybrid model that is at the required speed and accuracy for a given manipulation task.

We do this by using a coarse physics model to obtain a rough initial guess of the state at each time point of  a trajectory. Then, we evaluate the fine physics model in parallel across time starting from 
the initial guesses. Thereafter, we combine the coarse and fine predictions using the iterative Parareal algorithm \citep{parareal_Martin,parareal_Lions, parareal_Ruprecht_CoRR2015}. 

In this paper, we use this approach to perform physics-based predictions within a planner for robotic manipulation.  
Specifically, we consider the task of pushing an object to a goal location while avoiding obstacles.  
We provide a cheap coarse model and combine it with the Mujoco physics engine as the fine model. \footnote{We use Mujoco since it is recently the most widely used physics-engine for model-based planning 
\citep{Rajeswaran-RSS-18,pushing_fast_and_slow_Agboh_WAFR18,Matas-CORL2018,Ebert-CORL2018}. } 
The planner performs trajectory optimization to generate a control sequence, executes the first control in the sequence, and then re-runs the trajectory optimizer, in a model-predictive-control fashion.
We present this planner in Sec.~\ref{sec:planner}.  
As a baseline, we use the same planner, with the fine model, Mujoco, as the predictive model. 
We conduct experiments in simulation and on a real setup and show that the planner with hybrid physics models achieves the same success rates but faster.

To the best of our knowledge, the use 
of Parareal for contact dynamics (and in general for robotic planning and
control) has not been investigated before.  When used for contact dynamics, the
original Parareal formulation can produce infeasible states where rigid bodies
penetrate. We extend Parareal to handle these infeasible state updates through
projections to the feasible state space. 

\subsection{Related work}
Parareal has been used in different areas, e.g. to simulate incompressible laminar flows \citep{parareal_fluidsim}, or to simulate dynamics in quantum
chemistry \citep{parareal_quantum_control}.
It was introduced by \citet{parareal_Lions}.

Combining different physics models for robotic manipulation has been the topic
of other recent research as well, even though the focus has \textit{not} been
improving prediction speed.  \citet{combining_learned_analytical_Kloss_C0RR_17}
focus on the question of accuracy and generalization in combined
neural-analytical models.  \citet{augmenting_physics_Ajay_IROS18} focus on
modeling of the inherent stochastic nature of the real world physics, by combining an analytical, deterministic rigid-body simulator with a stochastic neural network. 

We can make physics engines faster by using larger simulation time steps,
however this decreases the accuracy and can quickly result in unstable
behavior.  To generate stable behaviour at large time-step sizes,
\citet{position_based_collocation_Pan_WAFR18} propose an integrator for
articulated body dynamics by using only position variables to formulate the
dynamic equation. Moreover,  \citet{Variational_Integrators_Fan_WAFR18} propose
linear-time variational integrators of arbitrarily high order for robotic
simulation and use them in trajectory optimization to complete robotics tasks.
Recent work have focused on making the underlying planning and control
algorithms faster. For example, \citet{GNMS_Giftthaler_CoRR17} introduced a
multiple-shooting variant of the trajectory optimizer - iterative linear
quadratic regulator (ilqr) which has shown impressive results for real-time
nonlinear optimal control of complex robotic systems
\citep{whole_body_mpc_buchli_RAL18, performance_DDP_Plancher_WAFR18}. 

\section{Combining physics models for planning}\label{sec:parareal}

\begin{figure*}[t]
\centering 
\captionsetup{justification=centering}
  \begin{subfigure}[b]{0.325\textwidth}
    \includegraphics[scale=0.27, angle=0]{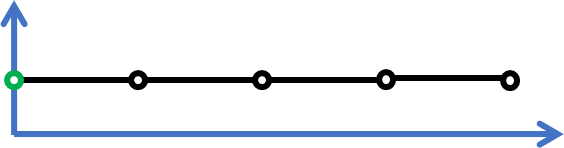}
    \begin{picture}(0,0)
	        \put(7,40){$\vec{x}$}
	        \put(109,17){n}
	        \put(2,5){0}
	        \put(25.5,5){1}
	        \put(51,5){2}
	        \put(77,5){3}
	        \put(102,5){4}
	\end{picture}
    %\vspace{1mm}
    \caption{Initial coarse physics predictions with $C$}
    \label{fig:coarse_initial_guess}
  \end{subfigure}
    \vspace{5mm}
    %%%
    \begin{subfigure}[b]{0.325\textwidth}
    \includegraphics[scale=0.27, angle=0]{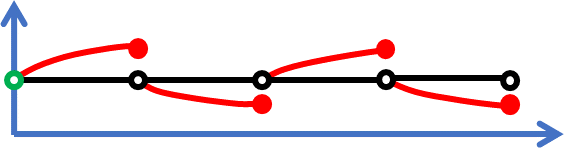}
    \begin{picture}(0,0)
	        \put(7,40){$\vec{x}$}
	        \put(109,17){n}
	        \put(2,5){0}
	        \put(25.5,5){1}
	        \put(51,5){2}
	        \put(77,5){3}
	        \put(102,5){4}
	\end{picture}
     %\vspace{1mm}
    \caption{Fine physics predictions with $F$ in \textit{parallel}}
    \label{fig:fine_predictions_1}
  \end{subfigure}
  %%%
    \begin{subfigure}[b]{0.325\textwidth}
    \includegraphics[scale=0.27, angle=0]{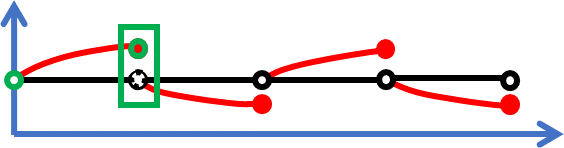}
    \begin{picture}(0,0)
	        \put(7,40){$\vec{x}$}
	        \put(109,17){n}
	        \put(2,5){0}
	        \put(25.5,5){1}
	        \put(51,5){2}
	        \put(77,5){3}
	        \put(102,5){4}
	\end{picture}
    %\vspace{1mm}
    \caption{Parareal update at time $n=1$}
    \label{fig:update_1}
  \end{subfigure}
    \begin{subfigure}[b]{0.325\textwidth}
    \includegraphics[scale=0.27, angle=0]{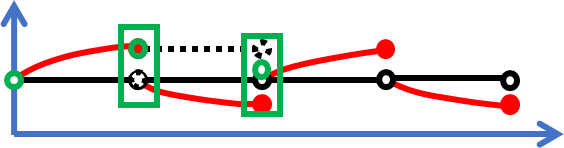}
    \begin{picture}(0,0)
	        \put(7,40){$\vec{x}$}
	        \put(109,17){n}
	        \put(2,5){0}
	        \put(25.5,5){1}
	        \put(51,5){2}
	        \put(77,5){3}
	        \put(102,5){4}
	\end{picture}
    \caption{Parareal update at time $n=2$}
    \label{fig:update_2}
  \end{subfigure}
  %%%
    \begin{subfigure}[b]{0.325\textwidth}
    \includegraphics[scale=0.27, angle=0]{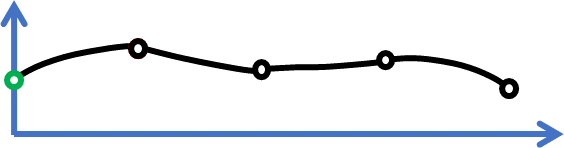}
    \begin{picture}(0,0)
	        \put(7,40){$\vec{x}$}
	        \put(109,17){n}
	        \put(2,5){0}
	        \put(25.5,5){1}
	        \put(51,5){2}
	        \put(77,5){3}
	        \put(102,5){4}
	\end{picture}
    \caption{Updates after 1 Parareal iteration}
    \label{fig:updated_initial_gueses }
  \end{subfigure}
    \begin{subfigure}[b]{0.325\textwidth}
    \includegraphics[scale=0.27, angle=0]{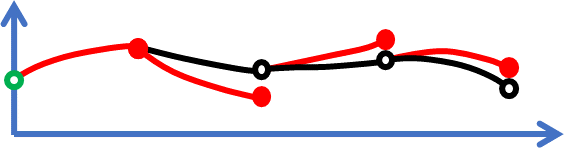}
    \begin{picture}(0,0)
	        \put(7,40){$\vec{x}$}
	        \put(109,17){n}
	        \put(2,5){0}
	        \put(25.5,5){1}
	        \put(51,5){2}
	        \put(77,5){3}
	        \put(102,5){4}
	\end{picture}
    \caption{Fine physics predictions for Parareal iteration 2 }
    \label{fig:fine_predictions_2}
  \end{subfigure}
%   \caption{The Parareal algorithm: (a) Initial coarse physics predictions across time with a cheap model ($C$), (b) Fine physics predictions in \textit{parallel} starting from coarse initial guesses with $F$. (c) A Parareal update at time $n=1$ as a linear combination of coarse and fine approximations of the state (d) A Parareal update at time $n=2$ using the updated state at time $n=1$, (e) Final trajectory updates after $k=1$ Parareal iteration, (f) Next Parareal iteration begins with fine physics predictions in \textit{parallel}.}
\caption{Combining coarse and fine physics with the Parareal algorithm}
  \label{fig:parareal_algorithm}
   %\vspace{-5mm}
 \end{figure*}
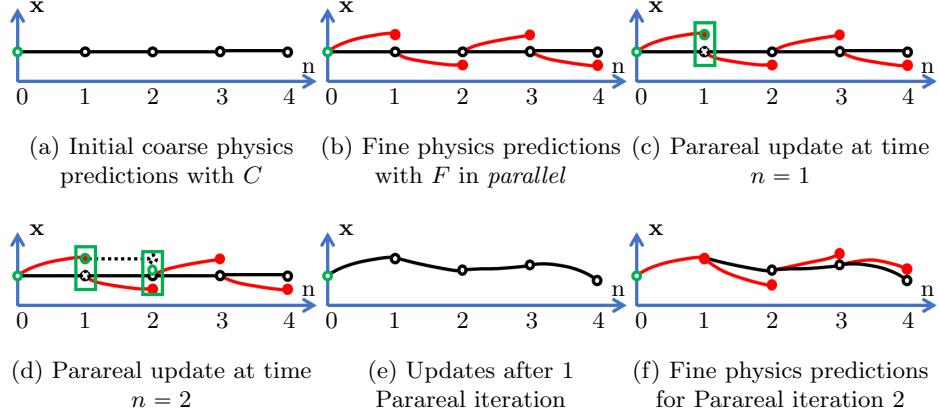
 %%%%%%%%%%%%%%%%%%%%%%%%%%%%%%%%%%%%%%%%%%%%%%%%%%%%%%%%%%%%%%%%%%%%%%%
Given an initial state $\vec{x}_0$ and a sequence of $N$ controls $\{\vec{u}_{0}, \vec{u}_{1}, \dots, \vec{u}_{N-1} \}$, we are interested in predicting the resulting sequence of states ${\{\vec{x}_{1}, \vec{x}_{2}, \dots, \vec{x}_{N} \}}$ of a physical system.  As an example, we consider the problem of pushing an object to a goal location with a cylindrical pusher. The system's state consists of the pose $\vec{q}$ and velocity $\dot{\vec{q}}$ of the pusher $P$ and slider $S$:
$
    \vec{x}_{n} = [\vec{q}^{P}_{n}, \vec{q}^{S}_{n}, \dot{\vec{q}}^{P}_{n}, \dot{\vec{q}}^{S}_{n}]
$.
The slider's pose consists of the translation and rotation of the object on the plane ${\vec{q}^{S}=[q^{S_x},q^{S_y},q^{S_{\theta}}]^T}$. The pusher's pose is: ${\vec{q}^{P}=[q^{P_x},q^{P_y}]^T}$ and control inputs are velocities ${\vec{u}_{n} = [u^{x}_{n}, u^{y}_{n}]^T}$ applied on the pusher for a  control duration of $\Delta t$.

To predict the next state of the system given an initial state and a control input, we need a physics model $F$. We use a general physics engine~\citep{mujoco} to model the system dynamics. It solves Newton's equations of motion for the complex multi-contact dynamics problem:
%%%
\vspace{-2mm}
\begin{equation}
    \label{eq:dynamics}
    \vec{x}_{n+1} = F(\vec{x}_n,\vec{u}_n,\Delta t).
    \vspace{-2mm}
\end{equation}
%%%

Normally, computing all states $\vec{x}_n$ happens in a serial fashion, by evaluating~\eqref{eq:dynamics} first for $n=0$, then for $n=1$, etc.
Instead, we replace this inherently serial procedure by a parallel-in-time integration process. Specifically, we adapt the Parareal algorithm for the contact-based manipulation problem. 

Parareal begins with a rough initial guess of the state at each time point $n$ of the trajectory as shown in Fig.~\ref{fig:coarse_initial_guess}. To get an initial guess, we define a second, coarse physics model:
%%%
\begin{equation}
    \label{eq:coarse_dynamics}
    \vec{x}_{n+1} = C(\vec{x}_n,\vec{u}_n,\Delta t)
    \vspace{-2mm}
\end{equation}
%%%
It needs to be computationally cheap relative to the fine model but does not need to be very accurate. 
%It is typically referred to as the coarse propagator (in contrast to the fine propagator $F$).

The next step is to evaluate the fine physics model in \textit{parallel} starting from $N$ initial guesses as shown in Fig.~\ref{fig:fine_predictions_1}. Thereafter, we do a coarse sweep across the time points. We start from the initial state $\vec{x}_0$ and make a coarse prediction for the next state $\vec{x}_1$ (dotted lines in Fig.~\ref{fig:update_1}). Now, we have 3 approximations of the state at time $n=1$. We linearly combine these approximations to get an update for $\vec{x}_{1}$ (in green). Then starting from this new update for $\vec{x}_1$, we make a coarse prediction for the next state $\vec{x}_2$ (dotted lines in Fig.~\ref{fig:update_2} at $n=2$). We combine the three approximations to get an update at $n=2$. We continue this coarse sweep for all time points to get the updated trajectory in Fig.~\ref{fig:updated_initial_gueses }. This is the end of the first iteration. We then repeat the whole process iteratively using new updates as initial guesses as shown in Fig.~\ref{fig:fine_predictions_2}.    

In summary, Parareal starts by computing rough initial guesses $\vec{x}^{k=0}_n$ of the system states using the coarse model. The newly introduced superscript $k$ counts the number of Parareal iterations. In each Parareal iteration, the guess is then refined via
\begin{equation}
    \label{eq:parareal}
    \vec{x}^{k+1}_{n+1} = C(\vec{x}^{k+1}_n, \vec{u}_n, \Delta t) + F(\vec{x}^{k}_n, \vec{u}_n, \Delta t) - C(\vec{x}^k_n, \vec{u}_n, \Delta t),
\end{equation}
for all timesteps ${n=0,\dots,N-1}$.
The key point in iteration~\eqref{eq:parareal} is that evaluating the fine physics model can be done in parallel for all ${n=0, \ldots, N-1}$, while only the fast coarse model has to be computed serially.

Parareal iterations converge exactly to the fine physics solution after $k=N$ iterations. After one iteration, $\vec{x}^{1}_{1}$ is exactly the fine solution. We can see this in Fig.~\ref{fig:update_1} where the two coarse physics predictions in Eq.~\ref{eq:parareal} are same and cancel out. After two iterations, $\vec{x}^{1}_{1}$ and $\vec{x}^{2}_{2}$ are exactly the fine solutions and so on. 

The idea is to stop Parareal at much earlier iterations such that it requires significantly less wall-clock time than running $F$ serially step-by-step. To do this, $C$  must be computationally much cheaper than $F$.

Parareal can be thought of as producing a spectrum of solutions increasing in
accuracy and computational cost, from the cheap coarse physics model to the
expensive fine physics model --- i.e. the $N$ different approximations after each iteration. 
An important question is which of these models to choose; i.e. how
many iterations of Parareal to use?  To decide on the required prediction
accuracy, we rely on recent work which analyzes the stochasticity in real-world
pushing \citep{million_ways_to_push,probabilistic_pushing_Bauza_ICRA17}. We
propose to stop Parareal when the approximation error with respect to the fine
model is below the real-world pushing stochasticity.

Note that, for the sake of simplicity, we assume here that the number of controls $N$ and the number of processors used to parallelize in time are identical.
% This need not be so.

% In case there are more controls than processors, we could combine multiple ${F(\vec{x}_n, \vec{u}_n, \Delta t})$ and parallelize over those aggregated updates.

%
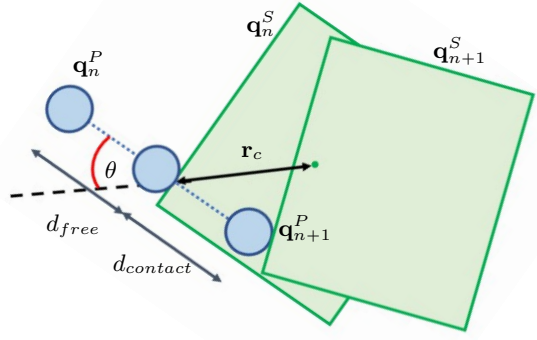
\begin{figure}[t]
\centering
\vspace{-17mm}
		\includegraphics[scale=0.5, angle=-125]{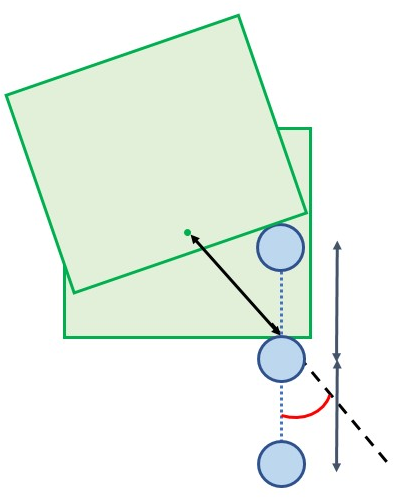} 
		\begin{picture}(0,0)
	        \put(-193,-120){$\theta$}
	        \put(-215,-140){$d_{free}$}
	        \put(-190,-155){$d_{contact}$}
	        \put(-128,-142){$\vec{q}^{P}_{n+1}$}
	        \put(-205,-80){$\vec{q}^{P}_{n}$}
	        \put(-140,-65){$\vec{q}^{S}_{n}$}
	        \put(-70,-75){$\vec{q}^{S}_{n+1}$}
	        \put(-142,-112){$\vec{r}_c$}
		\end{picture}
		\vspace{-17mm}
		\caption{Coarse physics model}
		\label{fig:coarse_physics_model}
		\vspace{-4mm}
\end{figure}
\vspace{-2mm}
\subsection{Expected speedup performance of Parareal}\label{sec:parareal_theoretical_speedup}
\vspace{-1mm}
We can describe the expected performance of Parareal by a simple theoretical model \citep{parareal_speedup_minion2010}. 
Let $c_{c}$ and $c_{f}$ be the time needed to compute the coarse physics model $C$ and the fine physics model $F$ respectively, for a duration of $\Delta t$.
% Then, running Parareal with $K$ iterations requires a wall-clock time $c_p$ of approximately
% \begin{equation}
% \label{eq:parareal_expected_time}
%     c_p = (1 + K) \cdot N \cdot c_c + K \cdot c_f
% \end{equation}
% neglecting overhead.
% By contrast, running the fine physics model serially takes wall-clock time $N c_f$.
The speedup of Parareal $s_p$ over the serial fine model is approximately:
\begin{align}
s_{p} = \dfrac{N\cdot c_{f}}{(1+K)\cdot N \cdot c_{c} + K\cdot c_{f}} = \dfrac{1}{(1+K)\frac{c_c}{c_f} + \frac{K}{N}}
%\label{eq:parareal_speedup}
\label{eq:parareal_expected_time}
\vspace{-5mm}
\end{align}
This illustrates the importance of finding a cheap coarse model that minimizes the ratio $c_{c}/c_{f}$. 
In that case, speedup will be determined mainly by the number of iterations $K$. For example, for a coarse model with negligible cost, after $K=1$ Parareal iteration with $N=4$ sub-intervals, then the theoretical speedup would be $s_{p} \sim N/K = 4$. 
That is, we can expect to make physics predictions about \textit{four times faster} than using only the fine physics model in serial.
\subsection{Coarse physics model}
As a case study, we consider the challenging problem of pushing an object. We seek a general coarse physics model with the following requirements: 

\begin{itemize}
    \item It must be significantly cheaper to compute with respect to the fine model. 
    \item It must provide a physics prediction for all possible pusher motions but can be inaccurate. 
    \item It must provide a prediction for sliders of any shape and inertial parameters. 
\end{itemize}{}

Instead of solving Newton's equation of motion for the multi-contact dynamics problem, we propose a simple kinematic pushing model $C(\vec{x}_{n}, \vec{u}_{n}, \Delta t)$. It moves the slider with the same linear velocity as the pusher, as long as there is contact between the two. We also apply a rotation to the slider, based on the position and direction of the contact, with respect to the center of the object.
Formally, given the linear velocity of the pusher as the controls ${\vec{u}_{n} = [u^{x}_{n}, u^{y}_{n}]^T}$, the next state of the system is given by;
\begin{align}\label{eq:coarse_slider_pose_update}
     \vec{q}^{S}_{n+1} = \vec{q}^{S}_{n} + [u^x_n, u^y_n, \omega]^T \cdot p_c \cdot \Delta t 
\end{align}
\vspace{-4mm}
\begin{align}
    p_{c} = \frac{d_{contact}}{d_{contact} + d_{free}}, \hspace{5mm} \omega = K_{\omega} \cdot \frac{||\vec{u}_{n}|| \cdot \sin{\theta}}{||\vec{r}_{c}||}
\end{align}
\vspace{-4mm}
\begin{align}\label{eq:coarse_slider_vel_update}
     \dot{\vec{q}}^{S}_{n+1} = \{ [u^x_n, u^y_n, \omega]^T
     \hspace{2mm} if \hspace{2mm} p_c > 0, \hspace{2mm} \dot{\vec{q}}^{S}_{n} \hspace{2mm} otherwise \}
\end{align}
\vspace{-4mm}
\begin{align}\label{eq:coarse_pusher_update}
    \vec{q}^{P}_{n+1} = \vec{q}^{P}_{n} + \vec{u}_{n} \cdot \Delta t, \hspace{5mm} \dot{\vec{q}}^{P}_{n+1} = \vec{u}_{n}.
\end{align}
In Eq.~\ref{eq:coarse_slider_pose_update} the slider's pose is updated as described above. Here, $p_{c}$ is the ratio of the distance $d_{contact}$ travelled by the pusher when in contact with the slider and the total pushing distance as shown in Fig.~\ref{fig:coarse_physics_model}. $\vec{r}_c$ is a vector from the contact point to the object's center (green dot) at the current state $\vec{q}^{S}_{n}$,  $\theta$ is the angle between the pushing direction and the vector $\vec{r}_c$. Moreover, $\omega$ is the coarse angular velocity induced by the pusher on the slider, where $K_{\omega}$ is a positive constant parameter. 

Also note that, even though Fig.~\ref{fig:coarse_physics_model} shows the pusher and slider in contact at the next time step, this does not have to be so; i.e. the coarse model can leave the two in separation. 

In Eq.~\ref{eq:coarse_slider_vel_update} the velocity of the slider is updated to be the same as the current pusher velocity if there is any contact. In Eq.~\ref{eq:coarse_pusher_update} the pusher position and velocity are updated. 

% Note that, even though the velocity terms here may seem unnecessary to update within the coarse model itself, their values are in fact used to initialize the fine model when the coarse predictions are combined with the fine predictions (as in Eq.~\ref{eq:parareal}) within Parareal.

\subsection{Infeasible states}
The new iterate $\vec{x}^{k+1}_{n+1}$ given by the Parareal iteration (Eq.~\ref{eq:parareal}) can be an infeasible state where the pusher and slider penetrate each other.
Contact dynamics is not well-defined for such states. It can lead to infinitely large object accelerations and an unstable fine physics model. We have not encountered such a problem of infeasible (or unallowed) states in other dynamics domains that Parareal has been applied to. 

To handle these cases, we project the infeasible state to the nearest feasible state. We write the following optimization problem:
\begin{align}
\vec{q}^{S*}_{n+1} = \argmin_{\vec{q}^{S}_{n+1}} ||\vec{q}^{S}_{n+1} - \vec{q}^{S_{infeasible}}_{n+1} ||, \hspace{3mm} s.t. \;\; d_p \leq 0
\end{align}

where $\vec{q}^{S_{infeasible}}_{n+1}$ is the infeasible slider's pose, and  $d_{p}$ is the penetration depth. The goal is to find the nearest slider pose $\vec{q}^{S*}_{n+1}$ that satisfies the no-penetration constraint $ d_p \leq 0$. 

We can use an off-the-shelf optimizer to find a solution rather efficiently. However, for simple systems we can analytically find the penetration depth and move the slider along the contact normal to resolve penetration.
%
%\vspace{3mm} 
%

In Sec.~\ref{sec:open_loop_pushing}, we evaluate the open-loop pushing performance of our hybrid physics models. Our goal is to use these hybrid models for planning and control. 
%%%%%%%%%%%%%%%%%%%%%%%%%%%%%%%%%%%%%%%%%%%%%%%%%%%%%%%%%%%%%%%%%%%%%%%%%%%%%%%%%%%%
\section{Push Planning and Control}\label{sec:planner}
\begin{figure}[t]
\centering
\captionsetup{justification=centering}
	\includegraphics[scale=0.5]{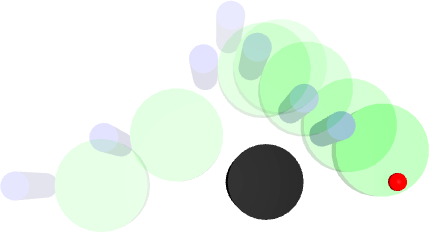} 
		\caption{Push planning with a hybrid physics model to avoid an obstacle (in black) while pushing the cylindrical slider to a goal location (in red).}
		\label{fig:obstacle_avoidance}
	\vspace{-5mm}
\end{figure}

\vspace{-1mm}
We use the predictive models described above in a planning and control framework for pushing an object on a table to a goal location, avoiding a static obstacle. This task retains many challenges of general robotic control through contact such as impulsive contact forces, under-actuation and hybrid dynamics (separation, sticking, sliding, e.t.c.). We present an example scene and execution in Fig.~\ref{fig:obstacle_avoidance}. 

To solve this problem, we take an optimization approach. Given the obstacle and goal position and geometry, the current state of the pusher and slider $\vec{x}_0$, and an initial candidate sequence of controls ${\{\vec{u}_0,\vec{u}_1,\dots,\vec{u}_{N-1}\}}$, the optimization procedure outputs an optimal sequence ${\{\vec{u}^*_0,\vec{u}^*_1,\dots,\vec{u}^*_{N-1}\}}$ according to some defined cost. We explain this optimization process, and the cost formulation that is optimized, below (Sec.~\ref{sec:optimization}). 

The predictive models that we have developed earlier in the paper are used within this optimizer to \textit{roll-out} a sequence of controls, to predict the states ${\{\vec{x}_1, \dots, \vec{x}_{N}\}}$, which are then used to compute the cost associated with those controls. 

Once the optimization produces a sequence of controls, we use it in a model-predictive-control (MPC) fashion, by executing only the first control in the sequence. Afterwards, we update $\vec{x}_0$ with the observed state of the system, and repeat the optimization to generate a new control sequence. This is repeated until task completion. We consider the task completed if the slider reaches the goal region (success), if it hits the obstacle (failure), if it falls off the edge of the table (failure) or if a maximum number of controls are executed before failure occurs.

When we repeat the optimization within MPC, we warm-start it by using the previously optimized control sequence as the initial candidate sequence. For the very first optimization, the initial candidate sequence is generated as a straight line push towards the goal (which collides with the obstacle in all our scenes).

Such an optimization-based MPC approach to pushing manipulation is frequently used \citep{mppi_push,pusher_slider,combining_learned_analytical_Kloss_C0RR_17,real_time_manipulation_Agboh_Humanoids18}. Here, our focus is to evaluate the performance of different predictive physics models described before in the paper within such a framework.

\subsection{Trajectory Optimization}\label{sec:optimization}
%%%
%%%
In this section we use the shorthand $\vec{u}_{0:N-1}$ to refer to the control sequence ${\{\vec{u}_0,\vec{u}_1,\dots,\vec{u}_{N-1}\}}$. Similarly for states we use $\vec{x}_{0:N}$.

Our goal is to find an optimal control sequence $\vec{u}^*_{0:N-1}$ for a planning horizon $N$, given an initial state $\vec{x}_{0}$, and an initial candidate control sequence $\vec{u}_{0:N-1}$.

We define the cost function $J$, for a given control sequence and the corresponding state sequence:
\vspace{-3mm}
\begin{align}
    J(\vec{x}_{0:N},\vec{u}_{0:N-1}) = \sum^{N-1}_{n=1} J_n(\vec{x}_{n}, \vec{u}_{n-1}, \vec{u}_{n}) + w \cdot J_{N}(\vec{x}_{N})), \label{eq:cost_function} 
    \vspace{-5mm}
\end{align}
where $J_n$ is the running cost at each step, $w$ is a positive weighting constant, $J_{N}$ is the terminal (final) cost function. 

The output of optimization is the minimizing control sequence:
\begin{align}
    \hspace{-2mm} \vec{u}^*_{0:N-1} = \argmin_{\vec{u}_{0:N-1}} J(\vec{x}_{0:N},\vec{u}_{0:N-1}), \hspace{2mm}
    s.t. \;\; \vec{x}_{n+1} = f(\vec{x}_{n}, \vec{u}_{n}, \Delta t),
    \label{eq:dynamics_constraint} \hspace{2mm}
    \vec{x}_0 \;\; fixed.
\end{align}
Here, $f$ is the system dynamics constraint that must be satisfied at all times.

We define the running cost for our pushing around an obstacle problem as:
\begin{align*}
    J_n(\vec{x}_{n}, \vec{u}_{n-1}, \vec{u}_{n}) = w_{s}\cdot (1/|| [q^{s_{x}}, q^{s_{y}}]^{T} - \vec{p}_{obs} ||^{2})  +  w_{p}\cdot (1/|| \vec{q}^{p} - \vec{p}_{obs} ||^{2}) \\ + w_{u}\cdot || \vec{u}_{t} - \vec{u}_{t-1}||^{2} + W_{E}
\end{align*}
where $w_{s}, w_{p}, w_{u}$ are positive constant weights. $\vec{p}_{obs}$ is the position vector of the static obstacle to be avoided, and $[q^{s_{x}}, q^{s_{y}}]$ are the x,y positions of the slider respectively. The above formula associates high cost for the slider or the pusher to approach the obstacle. It has a smoothness cost, to prevent high accelerations. Additionally, it has a constant edge cost $W_E$ for the slider falling off the table. 

We define the final cost $J_{N}$ as: $ J_{N}(\vec{x}_{n}) = || [q^{s_{x}}, q^{s_{y}}]^{T} - \vec{p}_{goal} ||^{2}$,
where $\vec{p}_{goal}$ is the position vector of the target/goal location. 

There exists different optimization methods to solve this problem \citep{ilqr,kalakrishnan2011stomp,mppi,pusher_slider,real_time_manipulation_Agboh_Humanoids18}. The main difference lies in the way the cost gradient is computed for a given sequence of controls. For ease of implementation, here we use derivative-free stochastic sampling-based methods \citep{kalakrishnan2011stomp,mppi,real_time_manipulation_Agboh_Humanoids18}. Particularly, we use the algorithm presented in \citet{real_time_manipulation_Agboh_Humanoids18}. In each optimization iteration, to find the cost gradient at the current control sequence, these stochastic sampling methods generate multiple noisified versions of the current control sequence, they roll-out these noisy controls to find the cost associated with each one, and use these costs to compute a numerical gradient, which is then used to update the control sequence to minimize the cost. The roll-out of these noisy control sequences to compute the resulting states and the cost is where we use the physics models. 
\vspace{-2mm}
\subsection{Parareal and MPC}
The Parareal framework for generating hybrid physics models yields itself well to model-predictive control.
Recall from Fig.~\ref{fig:updated_initial_gueses } that after 1 Parareal iteration, physics prediction for the first state $\vec{x}_1$ is exactly the same as the fine model. This means planning is accurate at least for the first action for all our hybrid models. This aligns well with our MPC framework since we execute only the first action and then re-plan.
\vspace{-3mm}
\section{Experiments and Results} 
\vspace{-2mm}
In our experiments, we address three key issues. First, we investigate how fast Parareal converges to the fine physics solution  
for pushing tasks. Second, we investigate the open-loop pushing performance of different physics models and compare it with real-world data. Finally, we investigate how the different physics models generated by Parareal (at different iterations) can be used within a planning and control framework to complete non-prehensile manipulation tasks. 

The goal of Parareal for push planning is to simulate physics faster than any given fine physics model. Therefore, we must stop Parareal at much earlier iterations to achieve meaningful speedup. However, we must also understand how different Parareal's predictions are with respect to the fine physics predictions at different iterations. We investigate Parareal's convergence for specific pushing examples in Sec.~\ref{sec:parareal_convergence_specific_examples}. In addition, in Sec.~\ref{sec:parareal_speedup_specific_examples}, we measure the empirical speedup we get from Parareal, and compare it with the theoretical speedup that was presented in Sec.~\ref{sec:parareal_theoretical_speedup}.

During push planning and control, we must decide on a Parareal iteration that gives us an acceptable approximation to the fine solution. To this end, in Sec~\ref{sec:open_loop_pushing}, we conduct open-loop pushing experiments. We statistically investigate Parareal's approximation error with respect to the fine solution, for a particular number of Parareal iterations. To do that, we start from a large number of random initial states and apply different control sequences open-loop. We then analyze how these statistical approximation errors compare with the standard deviation of the real-world uncertainty during similar pushing tasks \citep{million_ways_to_push}. 

In Sec~\ref{sec:push_planning_experiment} we investigate the performance of different physics models produced by Parareal, when used within the MPC framework described in Sec.~\ref{sec:planner} to push an object to a goal region, avoiding an obstacle. We compare the success rates and total task completion times for the different physics models (the coarse model, Parareal at different iterations, and the fine model).  We perform these experiments on a real robot setup Sec~\ref{sec:push_plan_real_robot_experiments}. 

\subsection{Parareal convergence for pushing}
\subsubsection{Parareal convergence for specific pushing examples:}
\label{sec:parareal_convergence_specific_examples}

\begin{figure*}[t]
	\centering
	\begin{subfigure}[b]{0.48\textwidth}
		\includegraphics[scale=0.39]{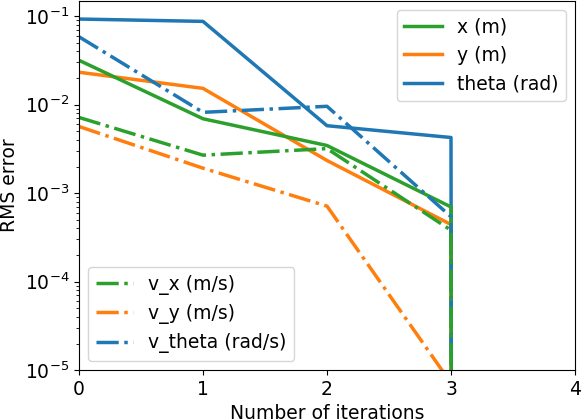} 
		\begin{picture}(0,0)
		\put(-38,25){
            \includegraphics[scale=0.14, angle=90]{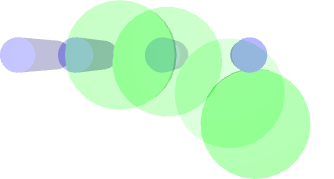} }
		\end{picture}
		\caption{Center cylinder push}
		\label{fig:center_cylinder_error}
	\end{subfigure}
	\begin{subfigure}[b]{0.48\textwidth}
		\includegraphics[scale=0.39]{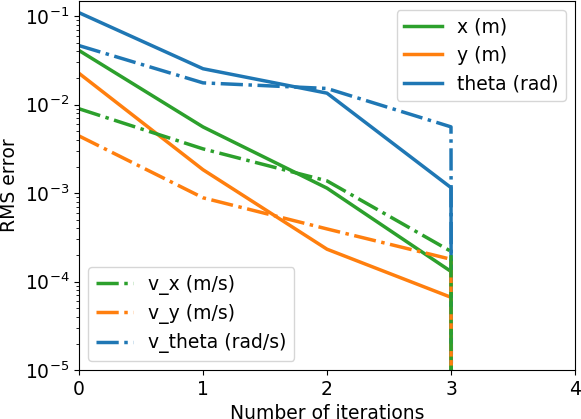}
		\begin{picture}(0,0)
		    \put(-42,25){
            \includegraphics[scale=0.14, angle=90]{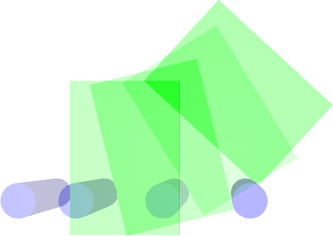} }
		\end{picture}
		\caption{Side box push}
		\label{fig:side_box_error}
	\end{subfigure}
	\caption{Root mean square error (in log scale) along the full trajectory for pushing a cylinder (a) and box (b) from the center and side respectively, for increasing Parareal iterations. The motions are illustrated lower-right in each plot.}
	\label{fig:parareal_convergence}
	\vspace{-7mm}
\end{figure*}

 We consider simulating the results of applying a control sequence starting from an initial state for a box and a cylinder as shown in Fig.~\ref{fig:parareal_convergence}. We consider four cases: pushing a cylinder from the center (Fig.~\ref{fig:center_cylinder_error}), pushing a cylinder from the side, pushing a box from the center, and finally pushing a box from the side (Fig.~\ref{fig:side_box_error}). The control sequence used here is $\vec{u}_{0:3} = \{[25, 0], [25, 0], [25, 0], [25, 0]\}mm/s$ where each control input is applied for a control duration $\Delta{t}=1s$ such that the total pushing distance is $100mm$.
 
We use the Mujoco \citep{mujoco} physics engine as the fine physics model. To make the fine model as fast as possible, we run it at the largest possible simulation time-step ($1 ms$) for our model. Beyond this time-step, the physics engine becomes unstable and breaks down. In addition, all experiments are run on a desktop computer (Intel(R) Xeon(R) CPU E3-1225 v3) with $N=4$ cores.
%%%
At each iteration of Parareal, we calculate the root mean square (RMS) error between Parareal's predictions and the physics engine's predictions of the corresponding sequence of states. These RMS errors can be seen in Fig.~\ref{fig:parareal_convergence} for two different cases. The results are similar for others. Note that the errors are given in log scale for the full state of the slider (pose and velocities). In general, we see a quick decrease in the error along the full trajectory starting from the large error of the coarse model at iteration 0. In addition, at the final iteration, we verify that the Parareal solution is exactly the same as using the fine model since the errors go to 0. 
\vspace{-3mm}
\subsubsection{Parareal speedup for specific pushing examples:}
\label{sec:parareal_speedup_specific_examples}
\begin{figure}[t]
\centering
	\includegraphics[scale=0.5]{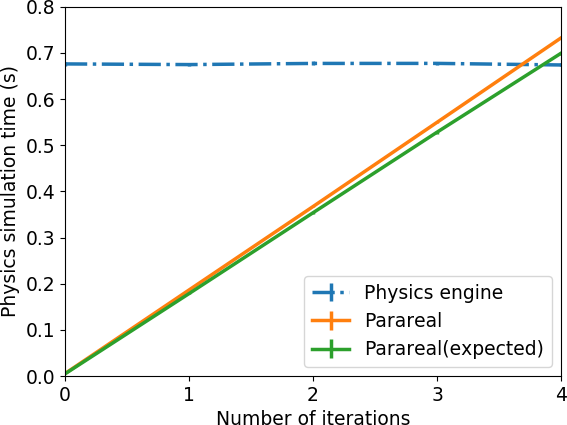} 
	\caption{Physics simulation time averaged over 100 runs for a box side push within 95 \% confidence interval of the mean.}
	\label{fig:parareal_time_plot}
	\vspace{-5mm}
\end{figure}

 Using each hybrid physics model, we repeatedly predict the sequence of states (which is deterministic for a given physics model) and record the total time it takes (which varies slightly depending on computer load). In  Fig.~\ref{fig:parareal_time_plot}, we see the average prediction time over 100 runs for each physics model for a box side push. These actual prediction times are close to the expected prediction time (Eq.~\ref{eq:parareal_expected_time}) for the different physics models. For example, at 1 Parareal iteration we spent 28\% of the time spent by the full physics engine, i.e. about \textit{four times faster}. The results for the cylinder side push, cylinder center push and box center push are similar to those in Fig.~\ref{fig:parareal_time_plot}.
\subsection{Open-loop pushing experiments}
\label{sec:open_loop_pushing}
\vspace{-2mm}
We compare the predictions of different physics models (Parareal iterations) for open-loop pushing. We start from 100 randomly sampled initial states\footnote{ We change the pusher's position along the direction perpendicular to the direction of motion. We sample uniformly within the edges of the slider (rectangle).}. At each initial state,  we used three different control sequences, giving 300 different slider trajectories. The three control sequences $\vec{u}^{\{1,2,3\}}$ that we used at each initial state were fixed, and are given by: 
\begin{align*}
    \vec{u}^{i}_{0:3} = v_{c} \cdot \{v,v,v,v\}, \hspace{3mm} v = [\cos(\alpha_{i}), \sin(\alpha_{i})], \hspace{3mm} \alpha_{i} = \{0^{\degree}, 15^{\degree}, -15^{\degree}\}, i \in {1,2,3}
\end{align*}
$v_c$ is a constant pushing speed:$v_{c} = 25mm/s$.   

We applied each control input for a duration of $\Delta{t} = 1.5s$ such that the total distance travelled by the pusher is $150 mm$ in all cases. We calculated for each physics model (Parareal iteration), the RMS difference of the final state (in comparison with the final state prediction of the fine physics model) for the 300 trajectories.

Our results are shown in Table.~\ref{table:physics_simulation_comparison}. We see that on the average the coarse physics model is quite inaccurate but with increasing Parareal iterations, the mean difference from the fine physics model goes to zero. However, to decide on how much error with respect to the physics engine is appropriate for pushing tasks, we look at the real-world's uncertainty for pushing dynamics. 

\citet{million_ways_to_push} provide real-world pushing data for a similar pusher-slider system. Starting at the same initial state, they push a box repeatedly in the real-world with a cylindrical pusher and record the resulting final positions. The pushing distance is $150mm$ with a quasi-static pushing speed of $20mm/s$. As shown in Table ~\ref{table:physics_simulation_comparison}, they record a translation standard deviation of $8.10mm$ and a rotation standard deviation of $4.20^{\degree}$ on a plywood surface. Notice that for Parareal after 2 iterations, we see a mean translation difference of $6.39mm$ and a mean rotation difference of $3.82^{\degree}$ when compared to the fine model (physics engine) predictions. 

We conclude that, for real-world purposes, it should not be necessary to run Parareal for more than 2 iterations, as approximating the physics engine more accurately than the inherent uncertainty in real-world pushing should not contribute to real-world performance. 
Note that, 2 Parareal iterations here corresponds to a) exactly the fine physics predictions for your first two actions and b) a model that is two times faster than the physics engine.
\begin{center}
\begin{table}[t]
\caption{Open-loop pushing} 
\centering 
\begin{tabular}{c c c} 
\hline\hline 
 & Mean trans. diff. (mm) & Mean rot. diff. (deg)\\ 
\hline \hline 
Coarse Physics  & 62.67  & 17.63 \\
Parareal-1 iter. & 28.43 & 6.30  \\
Parareal-2 iter. & 6.39 & 3.82  \\
Parareal-3 iter. & 2.47 & 0.79  \\
Parareal-4 iter. & 0.00 & 0.00 \\ 
\hline\hline
 & Trans. std. (mm) & Rot. std. (deg)\\
\hline\hline
Push dataset \citep{million_ways_to_push}  & 8.10 & 4.20 \\ 
\hline 
\end{tabular}
\label{table:physics_simulation_comparison}
\end{table}
\vspace{-7mm}
\end{center}
\vspace{-5mm}
\subsection{Push planning and control with hybrid physics models}
\label{sec:push_planning_experiment}
\label{sec:push_plan_real_robot_experiments}
We measure the performance of the different physics models when used within the optimization and MPC framework, described in Sec.~\ref{sec:planner}, to push an object to a goal region, while avoiding an obstacle.

\begin{figure*}[t]
	\centering
	\captionsetup{justification=centering}
    \begin{subfigure}[b]{0.23\textwidth}
	\centering
	\includegraphics[height=1.4in, scale=0.53]{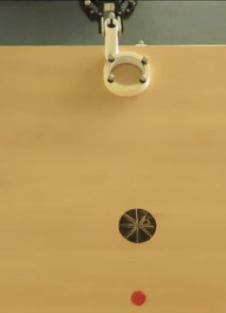} 
	\end{subfigure}
	\begin{subfigure}[b]{0.23\textwidth}
	\centering
	\hspace{-7mm}
		\includegraphics[height=1.4in, scale=0.53]{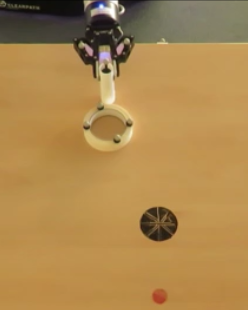}
	\end{subfigure}
	\begin{subfigure}[b]{0.23\textwidth}
	\centering
	\hspace{-10mm}
		\includegraphics[height=1.4in, scale=0.53]{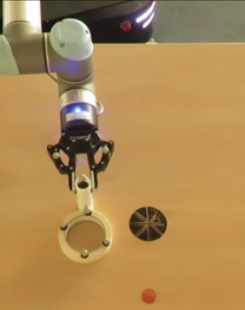}
	\end{subfigure}
	\begin{subfigure}[b]{0.23\textwidth}
	\centering
	\hspace{-14mm}
		\includegraphics[height=1.4in, scale=0.55]{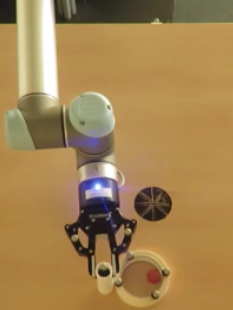}
	\end{subfigure}
	\caption{Pushing with a hybrid physics model. We complete the push planning task about \textbf{four times faster} than a physics engine.}
	\label{fig:real_robot_snapshots}
	\vspace{-5mm}
\end{figure*}

Our real robot setup is shown in Fig.~\ref{fig:real_robot_snapshots} where we have a Robotiq two-finger gripper holding the cylindrical pusher. We place markers on the pusher and slider to sense their full pose in the environment with the OptiTrack motion capture system.  We consider 5 randomly generated scenes (one shown in Fig.~\ref{fig:real_robot_snapshots}) where the pusher must avoid the obstacle at the center of the table before bringing the slider to the goal location. For each of the 5 scenes, we used the coarse physics model, Parareal iterations (1,2, and 3), and the full physics engine for push planning and control. That is a total of 25 planning and control runs with the real robot. 
We plan using our various physics models and execute actions in the real world in an MPC fashion. 
For the stochastic trajectory optimizer, our control sequences contain $4$ control inputs each applied for a control duration of $\Delta {t} = 1s$. In addition, we use $20$ noisy control sequence samples (as explained in Sec.~\ref{sec:optimization}) per optimization iteration for the trajectory optimizer with an exploration variance of $10^{-4}$. Normally, each noisy control sequence of the optimizer is rolled out independently. However, since we use a standard quad-core desktop PC, the \textit{parallelization is across time only}. 

As the pusher attempts to bring the slider to a desired goal location, there are three possible failure modes. First, we declare failure when the slider collides with the static obstacle. Second, we declare failure when the pusher is unable to bring the slider to the goal location after executing 20 actions (5 times the number of actions in a given control sequence). Third, we declare failure when the slider falls off the edge of the table. 

\begin{figure}[t]
\centering
	\includegraphics[scale=0.5]{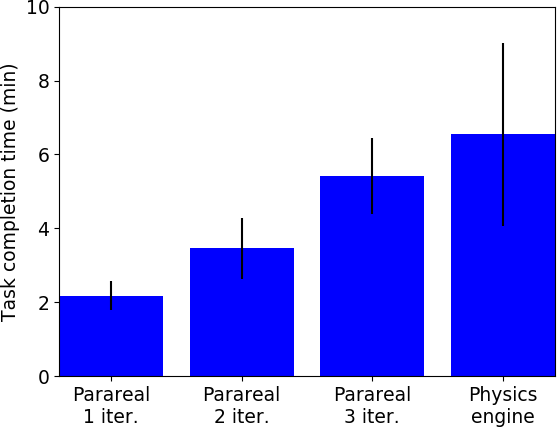}
\caption{Total task completion time (within 95 percent confidence interval of the mean) for push planning with obstacle avoidance using different physics models for 100 randomly sampled initial states.}
\label{fig:real_task_completion_time}
\vspace{-6mm}
\end{figure}

For each hybrid physics model, we achieved 100\% success rate on the real robot.  This is same as using a full physics engine, only significantly faster Fig.~\ref{fig:real_task_completion_time}.  For instance, using 1 Parareal iteration, we can complete the push planning task  about \textit{four times faster} than the physics engine. Note that the planning times here can be reduced by parallelizing the stochastic trajectory optimizer on a PC with more cores. Currently, the 4 cores of our PC is used to \textit{parallelize across time}.  
%%%
Furthermore, in all the 5 scenes considered, the robot was unable to complete the push planning task by using only the coarse model. We present snapshots from the experiments in Fig.~\ref{fig:real_robot_snapshots}. 
\vspace{-3mm}
\section{Discussion and Future Work}
\vspace{-3mm}
The method we presented here opens up important and exciting questions for manipulation planning and control. We discuss some of them briefly below.

\textit{Different coarse models}: We proposed a cheap, general and accurate enough coarse pushing model. This can be replaced with other models. 

We would like to explore learned pushing/poking models \citep{agrawal2016learning,finn2016unsupervised,finn2017deep,effect_aware_push,combining_learned_analytical_Kloss_C0RR_17,convex_polynomial_model_Zhou_IJRR18,augmenting_physics_Ajay_IROS18} as coarse models. Devising/learning coarse models for other more complex planning environments (e.g. for manipulation in clutter) is another important problem.

\textit{Degree of parallelization}: In this paper, we used 4 cores for parallelization. With a higher number of parallel slices, it may be possible to get higher speedups.

\textit{Task adaptivity}: Different tasks require different degree of accuracy \citep{pushing_fast_and_slow_Agboh_WAFR18}. For example, think of searching for a sock in the sock drawer, versus searching for a wine glass in the glass cabinet. It is okay for a robot's physics predictions to be coarse in the former example, which is not the case in the latter. Parareal can be used to explore this spectrum, and generate coarse predictions when it is sufficient for the task, and more accurate predictions as the task requires.
\vspace{-3mm}
\section{Acknowledgements}
\vspace{-3mm}
This project was funded from the European Unions's Horizon 2020 programme under the Marie Sklodowska Curie grant No. 746143, and from the UK EPSRC under grants EP/P019560/1, EP/R031193/1, and studentship 1879668.

\vspace{2mm}
\hspace{-5mm}
\bibliographystyle{plainnat}
{\let\clearpage\relax \bibliography{bibliography_file} }
\end{document}